\documentclass[11pt,a4paper]{article}
\usepackage[hyperref]{acl2018}
\usepackage{times}
\usepackage{latexsym}

\usepackage{url}

\usepackage{multirow} 
\usepackage{booktabs} 

\usepackage{amsmath} 
\newcommand{\argmax}[1]{\underset{#1}{\operatorname{arg}\,\operatorname{max}}\;}

\usepackage{graphicx} 

\aclfinalcopy 
\def\aclpaperid{1298} 

\title{A robust self-learning method for \\ fully unsupervised cross-lingual mappings of word embeddings}
\author{Mikel Artetxe \and Gorka Labaka \and Eneko Agirre \\
IXA NLP Group \\
University of the Basque Country (UPV/EHU) \\
\texttt{\{mikel.artetxe,gorka.labaka,e.agirre\}@ehu.eus} \\
}

\date{}

\begin{document}
\maketitle
\begin{abstract}
Recent work has managed to learn cross-lingual word embeddings without parallel data by mapping monolingual embeddings to a shared space through adversarial training. However, their evaluation has focused on favorable conditions, using comparable corpora or closely-related languages, and we show that they often fail in more realistic scenarios. This work proposes an alternative approach based on a fully unsupervised initialization that explicitly exploits the structural similarity of the embeddings, and a robust self-learning algorithm that iteratively improves this solution. Our method succeeds in all tested scenarios and obtains the best published results in standard datasets, even surpassing previous supervised systems. Our implementation is released as an open source project at \url{https://github.com/artetxem/vecmap}.
\end{abstract}

\section{Introduction} \label{sec:introduction}

Cross-lingual embedding mappings have shown to be an effective way to learn bilingual word embeddings \citep{mikolov2013exploiting,lazaridou2015hubness}. The underlying idea is to independently train the embeddings in different languages using monolingual corpora, and then map them to a shared space through a linear transformation. This allows to learn high-quality cross-lingual representations without expensive supervision, opening new research avenues like unsupervised neural machine translation \citep{artetxe2018unsupervised,lample2018unsupervised}.

While most embedding mapping methods rely on a small seed dictionary, adversarial training has recently produced exciting results in fully unsupervised settings \citep{zhang2017adversarial,zhang2017earth,conneau2018word}. However, their evaluation has focused on particularly favorable conditions, limited to closely-related languages or comparable Wikipedia corpora. When tested on more realistic scenarios, we find that they often fail to produce meaningful results. For instance, none of the existing methods works in the standard English-Finnish dataset from \citet{artetxe2017learning}, obtaining translation accuracies below 2\% in all cases (see Section \ref{sec:results}).

On another strand of work, \citet{artetxe2017learning} showed that an iterative self-learning method is able to bootstrap a high quality mapping from very small seed dictionaries (as little as 25 pairs of words). However, their analysis reveals that the self-learning method gets stuck in poor local optima when the initial solution is not good enough, thus failing for smaller training dictionaries.

\begin{figure*}[t] \centering
\includegraphics[width=0.9\textwidth]{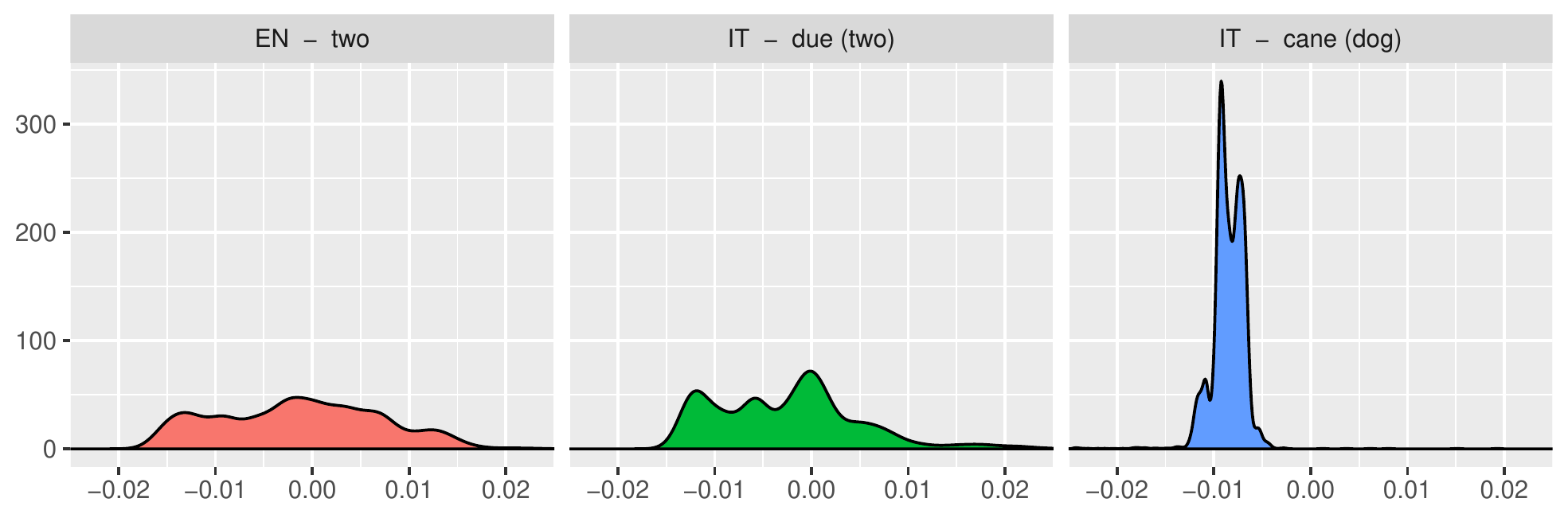}
\caption{Motivating example for our unsupervised initialization method, showing the similarity distributions of three words (corresponding to the smoothed density estimates from the normalized square root of the similarity matrices as defined in Section \ref{subsec:init}). Equivalent translations (two and \textit{due}) have more similar distributions than non-related words (two and \textit{cane} - meaning dog). This observation is used to build an initial solution that is later improved through self-learning.}
\label{fig:example-init}
\end{figure*}

In this paper, we follow this second approach and propose a new unsupervised method to build an initial solution without the need of a seed dictionary, based on the observation that, given the similarity matrix of all words in the vocabulary, each word has a different distribution of similarity values. Two equivalent words in different languages should have a similar distribution, and we can use this fact to induce the initial set of word pairings (see Figure \ref{fig:example-init}). We combine this initialization with a more robust self-learning method, which is able to start from the weak initial solution and iteratively improve the mapping. Coupled together, we provide a fully unsupervised cross-lingual mapping method that is effective in realistic settings, converges to a good solution in all cases tested, and sets a new state-of-the-art in bilingual lexicon extraction, even surpassing previous supervised methods.


\section{Related work} \label{sec:related_work}

Cross-lingual embedding mapping methods work by independently training word embeddings in two languages, and then mapping them to a shared space using a linear transformation.

Most of these methods are \textbf{supervised}, and use a bilingual dictionary of a few thousand entries to learn the mapping. Existing approaches can be classified into regression methods, which map the embeddings in one language using a least-squares objective \citep{mikolov2013exploiting,shigeto2015ridge,dinu2015improving}, canonical methods, which map the embeddings in both languages to a shared space using canonical correlation analysis and extensions of it \citep{faruqui2014improving,lu2015deep}, orthogonal methods, which map the embeddings in one or both languages under the constraint of the transformation being orthogonal \citep{xing2015normalized,artetxe2016learning,zhang2016ten,smith2017offline}, and margin methods, which map the embeddings in one language to maximize the margin between the correct translations and the rest of the candidates \citep{lazaridou2015hubness}. \citet{artetxe2018generalizing} showed that many of them could be generalized as part of a multi-step framework of linear transformations.

A related research line is to adapt these methods to the \textbf{semi-supervised} scenario, where the training dictionary is much smaller and used as part of a bootstrapping process. While similar ideas where already explored for traditional count-based vector space models \citep{peirsman2010crosslingual,vulic2013study}, \citet{artetxe2017learning} brought this approach to pre-trained low-dimensional word embeddings, which are more widely used nowadays. More concretely, they proposed a self-learning approach that alternates the mapping and dictionary induction steps iteratively, obtaining results that are comparable to those of supervised methods when starting with only 25 word pairs.

A practical approach for reducing the need of bilingual supervision is to design \textbf{heuristics to build the seed dictionary}. The role of the seed lexicon in learning cross-lingual embedding mappings is analyzed in depth by \newcite{vulic2016role}, who propose using document-aligned corpora to extract the training dictionary. A more common approach is to rely on shared words and cognates \cite{peirsman2010crosslingual,smith2017offline}, while \citet{artetxe2017learning} go further and restrict themselves to shared numerals. However, while these approaches are meant to eliminate the need of bilingual data in practice, they also make strong assumptions on the writing systems of languages (e.g. that they all use a common alphabet or Arabic numerals).
Closer to our work, a recent line of \textbf{fully unsupervised} approaches drops these assumptions completely, and attempts to learn cross-lingual embedding mappings based on distributional information alone. For that purpose, existing methods rely on adversarial training. This was first proposed by \citet{micelibarone2016towards}, who combine an encoder that maps source language embeddings into the target language, a decoder that reconstructs the source language embeddings from the mapped embeddings, and a discriminator that discriminates between the mapped embeddings and the true target language embeddings. Despite promising, they conclude that their model ``is not competitive with other cross-lingual representation approaches''. \citet{zhang2017adversarial} use a very similar architecture, but incorporate additional techniques like noise injection to aid training and report competitive results on bilingual lexicon extraction. \citet{conneau2018word} drop the reconstruction component, regularize the mapping to be orthogonal, and incorporate an iterative refinement process akin to self-learning, reporting very strong results on a large bilingual lexicon extraction dataset. Finally, \citet{zhang2017earth} adopt the earth mover's distance for training, optimized through a Wasserstein generative adversarial network followed by an alternating optimization procedure. However, all this previous work used comparable Wikipedia corpora in most experiments and, as shown in Section \ref{sec:results}, face difficulties in more challenging settings.

\section{Proposed method} \label{sec:method}

Let $X$ and $Z$ be the word embedding matrices in two languages, so that their $i$th row $X_{i*}$ and $Z_{i*}$ denote the embeddings of the $i$th word in their respective vocabularies. Our goal is to learn the linear transformation matrices $W_X$ and $W_Z$ so the mapped embeddings $XW_X$ and $ZW_Z$ are in the same cross-lingual space. At the same time, we aim to build a dictionary between both languages, encoded as a sparse matrix $D$ where $D_{ij} = 1$ if the $j$th word in the target language is a translation of the $i$th word in the source language.

Our proposed method consists of four sequential steps: a pre-processing that normalizes the embeddings (\S \ref{subsec:normalization}), a fully unsupervised initialization scheme that creates an initial solution (\S \ref{subsec:init}), a robust self-learning procedure that iteratively improves this solution (\S \ref{subsec:self-learning}), and a final refinement step that further improves the resulting mapping through symmetric re-weighting (\S \ref{subsec:refinement}).

\subsection{Embedding normalization} \label{subsec:normalization}

Our method starts with a pre-processing that length normalizes the embeddings, then mean centers each dimension, and then length normalizes them again. The first two steps have been shown to be beneficial in previous work \citep{artetxe2016learning}, while the second length normalization guarantees the final embeddings to have a unit length. As a result, the dot product of any two embeddings is equivalent to their cosine similarity and directly related to their Euclidean distance\footnote{Given two length normalized vectors $u$ and $v$, $u \cdot v = \cos(u, v) = 1 - ||u - v||^2/2$.}, and can be taken as a measure of their similarity.

\subsection{Fully unsupervised initialization} \label{subsec:init}

The underlying difficulty of the mapping problem in its unsupervised variant is that the word embedding matrices $X$ and $Z$ are unaligned across both axes: neither the $i$th vocabulary item $X_{i*}$ and $Z_{i*}$ nor the $j$th dimension of the embeddings $X_{*j}$ and $Z_{*j}$ are aligned, so there is no direct correspondence between both languages. In order to overcome this challenge and build an initial solution, we propose to first construct two alternative representations $X'$ and $Z'$ that are aligned across their $j$th dimension $X'_{*j}$ and $Z'_{*j}$, which can later be used to build an initial dictionary that aligns their respective vocabularies.

Our approach is based on a simple idea: while the axes of the original embeddings $X$ and $Z$ are different in nature, both axes of their corresponding similarity matrices $M_X = XX^T$ and $M_Z = ZZ^T$ correspond to words, which can be exploited to reduce the mismatch to a single axis. More concretely, assuming that the embedding spaces are perfectly isometric, the similarity matrices $M_X$ and $M_Z$ would be equivalent up to a permutation of their rows and columns, where the permutation in question defines the dictionary across both languages. In practice, the isometry requirement will not hold exactly, but it can be assumed to hold approximately, as the very same problem of mapping two embedding spaces without supervision would otherwise be hopeless. Based on that, one could try every possible permutation of row and column indices to find the best match between $M_X$ and $M_Z$, but the resulting combinatorial explosion makes this approach intractable.

In order to overcome this problem, we propose to first sort the values in each row of $M_X$ and $M_Z$, resulting in matrices $\operatorname{sorted}(M_X)$ and $\operatorname{sorted}(M_Z)$\footnote{Note that the values in each row are sorted independently from other rows.}. Under the strict isometry condition, equivalent words would get the exact same vector across languages, and thus, given a word and its row in $\operatorname{sorted}(M_X)$, one could apply nearest neighbor retrieval over the rows of $\operatorname{sorted}(M_Z)$ to find its corresponding translation.

On a final note, given the singular value decomposition $X = USV^T$, the similarity matrix is $M_X=US^2U^T$. As such, its square root $\sqrt{M_X} = USU^T$ is closer in nature to the original embeddings, and we also find it to work better in practice. We thus compute $\operatorname{sorted}(\sqrt{M_X})$ and $\operatorname{sorted}(\sqrt{M_Z})$ and normalize them as described in Section \ref{subsec:normalization}, yielding the two matrices $X'$ and $Z'$ that are later used to build the initial solution for self-learning (see Section \ref{subsec:self-learning}).

In practice, the isometry assumption is strong enough so the above procedure captures some cross-lingual signal. In our English-Italian experiments, the average cosine similarity across the gold standard translation pairs is 0.009 for a random solution, 0.582 for the optimal supervised solution, and 0.112 for the mapping resulting from this initialization. While the latter is far from being useful on its own (the accuracy of the resulting dictionary is only 0.52\%), it is substantially better than chance, and it works well as an initial solution for the self-learning method described next.

\subsection{Robust self-learning} \label{subsec:self-learning}

Previous work has shown that self-learning can learn high-quality bilingual embedding mappings starting with as little as 25 word pairs \citep{artetxe2017learning}. In this method, training iterates through the following two steps until convergence:
\begin{enumerate}
\item Compute the optimal orthogonal mapping maximizing the similarities for the current dictionary $D$:
\[\argmax{W_X, W_Z} \sum_i \sum_j D_{ij} ((X_{i*}W_X) \cdot (Z_{j*}W_Z)) \]
An optimal solution is given by $W_X=U$ and $W_Z=V$, where $USV^T = X^TDZ$ is the singular value decomposition of $X^TDZ$.
\item Compute the optimal dictionary over the similarity matrix of the mapped embeddings $XW_XW_Z^TZ^T$. This typically uses nearest neighbor retrieval from the source language into the target language, so $D_{ij} = 1$ if $j = \operatorname{argmax}_k \left( X_{i*} W_X \right) \cdot \left( Z_{k*} W_Z \right)$ and $D_{ij} = 0$ otherwise.
\end{enumerate}

The underlying optimization objective is independent from the initial dictionary, and the algorithm is guaranteed to converge to a local optimum of it. However, the method does not work if starting from a completely random solution, as it tends to get stuck in poor local optima in that case. For that reason, we use the unsupervised initialization procedure at Section \ref{subsec:init} to build an initial solution. However, simply plugging in both methods did not work in our preliminary experiments, as the quality of this initial method is not good enough to avoid poor local optima. For that reason, we next propose some key improvements in the dictionary induction step to make self-learning more robust and learn better mappings:
\begin{itemize}
\item \textbf{Stochastic dictionary induction}. In order to encourage a wider exploration of the search space, we make the dictionary induction stochastic by randomly keeping some elements in the similarity matrix with probability $p$ and setting the remaining ones to 0. As a consequence, the smaller the value of $p$ is, the more the induced dictionary will vary from iteration to iteration, thus enabling to escape poor local optima. So as to find a fine-grained solution once the algorithm gets into a good region, we increase this value during training akin to simulated annealing, starting with $p=0.1$ and doubling this value every time the objective function at step 1 above does not improve more than $\epsilon=10^{-6}$ for 50 iterations.
\item \textbf{Frequency-based vocabulary cutoff}. The size of the similarity matrix grows quadratically with respect to that of the vocabularies. This does not only increase the cost of computing it, but it also makes the number of possible solutions grow exponentially\footnote{There are $m^n$ possible combinations that go from a source vocabulary of $n$ entries to a target vocabulary of $m$ entries.}, presumably making the optimization problem harder. Given that less frequent words can be expected to be noisier, we propose to restrict the dictionary induction process to the $k$ most frequent words in each language, where we find $k=20,000$ to work well in practice.
\item \textbf{CSLS retrieval}. \citet{dinu2015improving} showed that nearest neighbor suffers from the hubness problem. This phenomenon is known to occur as an effect of the curse of dimensionality, and causes a few points (known as \textit{hubs}) to be nearest neighbors of many other points \citep{radovanovic2010hubs,radovanovic2010existence}. Among the existing solutions to penalize the similarity score of hubs, we adopt the Cross-domain Similarity Local Scaling (CSLS) from \citet{conneau2018word}. Given two mapped embeddings $x$ and $y$, the idea of CSLS is to compute $\operatorname{r_T}(x)$ and $\operatorname{r_S}(y)$, the average cosine similarity of $x$ and $y$ for their $k$ nearest neighbors in the other language, respectively. Having done that, the corrected score $\operatorname{CSLS}(x, y) = 2\cos(x, y) - \operatorname{r_T}(x) - \operatorname{r_S}(y)$. Following the authors, we set $k=10$.
\item \textbf{Bidirectional dictionary induction}. When the dictionary is induced from the source into the target language, not all target language words will be present in it, and some will occur multiple times. We argue that this might accentuate the problem of local optima, as repeated words might act as strong attractors from which it is difficult to escape. In order to mitigate this issue and encourage diversity, we propose inducing the dictionary in both directions and taking their corresponding concatenation, so $D = D_{X \rightarrow Z} + D_{Z \rightarrow X}$.
\end{itemize}

In order to build the \textbf{initial dictionary}, we compute $X'$ and $Z'$ as detailed in Section \ref{subsec:init} and apply the above procedure over them. As the only difference, this first solution does not use the stochastic zeroing in the similarity matrix, as there is no need to encourage diversity ($X'$ and $Z'$ are only used once), and the threshold for vocabulary cutoff is set to $k=4,000$, so $X'$ and $Z'$ can fit in memory. Having computed the initial dictionary, $X'$ and $Z'$ are discarded, and the remaining iterations are performed over the original embeddings $X$ and $Z$.

\subsection{Symmetric re-weighting} \label{subsec:refinement}

As part of their multi-step framework, \citet{artetxe2018generalizing} showed that re-weighting the target language embeddings according to the cross-correlation in each component greatly improved the quality of the induced dictionary. Given the singular value decomposition $USV^T = X^TDZ$, this is equivalent to taking $W_X=U$ and $W_Z=VS$, where $X$ and $Z$ are previously whitened applying the linear transformations $(X^TX)^{-\frac{1}{2}}$ and $(Z^TZ)^{-\frac{1}{2}}$, and later de-whitened applying $U^T (X^TX)^{\frac{1}{2}} U$ and $V^T (Z^TZ)^{\frac{1}{2}} V$.

However, re-weighting also accentuates the problem of local optima when incorporated into self-learning as, by increasing the relevance of dimensions that best match for the current solution, it discourages to explore other regions of the search space. For that reason, we propose using it as a final step once self-learning has converged to a good solution. Unlike \citet{artetxe2018generalizing}, we apply re-weighting symmetrically in both languages, taking $W_X=US^{\frac{1}{2}}$ and $W_Z=VS^{\frac{1}{2}}$. This approach is neutral in the direction of the mapping, and gives good results as shown in our experiments.

\begin{table*}[t]
\begin{small}
\begin{center}
  \addtolength{\tabcolsep}{-4pt}
  \begin{tabular}{lrrrrrrrrrrrrrr}
    \toprule
    & \multicolumn{4}{c}{\bf ES-EN} & & \multicolumn{4}{c}{\bf IT-EN} & & \multicolumn{4}{c}{\bf TR-EN} \\
    \cmidrule{2-5} \cmidrule{7-10} \cmidrule{12-15}
    & \multicolumn{1}{c}{\bf best} & \multicolumn{1}{c}{\bf avg} & \multicolumn{1}{c}{\bf s} & \multicolumn{1}{c}{\bf t} & & \multicolumn{1}{c}{\bf best} & \multicolumn{1}{c}{\bf avg} & \multicolumn{1}{c}{\bf s} & \multicolumn{1}{c}{\bf t} & & \multicolumn{1}{c}{\bf best} & \multicolumn{1}{c}{\bf avg} & \multicolumn{1}{c}{\bf s} & \multicolumn{1}{c}{\bf t} \\
    \midrule
    \citet{zhang2017adversarial}, $\lambda = 1$ & 71.43 & 68.18 & \bf 10 & 13.2 & & 60.38 & 56.45 & \bf 10 & 12.3 & & 0.00 & 0.00 & 0 & 13.0 \\
    \citet{zhang2017adversarial}, $\lambda = 10$ & 70.24 & 66.37 & \bf 10 & 13.0 & & 57.64 & 52.60 & \bf 10 & 12.6 & & 21.07 & 17.95 & \bf 10 & 13.2 \\
    \citet{conneau2018word}, code & 76.18 & 75.82 & \bf 10 & 25.1 & & \bf 67.32 & \bf 67.00 & \bf 10 & 25.9 & & 32.64 & 14.34 & 5 & 25.3 \\
    \citet{conneau2018word}, paper & 76.15 & 75.81 & \bf 10 & 25.1 & & 67.21 & 60.22 & 9 & 25.5 & & 29.79 & 16.48 & 7 & 25.5 \\
    Proposed method & \bf 76.43 & \bf 76.28 & \bf 10 & \bf 0.6 & & 66.96 & 66.92 & \bf 10 & \bf 0.9 & & \bf 36.10 & \bf 35.93 & \bf 10 & \bf 1.7 \\
    \bottomrule
  \end{tabular}
\end{center}
\end{small}
\caption{Results of unsupervised methods on the dataset of \citet{zhang2017adversarial}. We perform 10 runs for each method and report the best and average accuracies (\%), the number of successful runs (those with \textgreater 5\% accuracy) and the average runtime (minutes).}
\label{tab:results_zhang}
\end{table*}

\begin{table*}[t]
\begin{small}
\begin{center}
  \addtolength{\tabcolsep}{-4pt}
  \begin{tabular}{lrrrrrrrrrrrrrrrrrrr}
    \toprule
    & \multicolumn{4}{c}{\bf EN-IT} & & \multicolumn{4}{c}{\bf EN-DE} & & \multicolumn{4}{c}{\bf EN-FI} & & \multicolumn{4}{c}{\bf EN-ES} \\
    \cmidrule{2-5} \cmidrule{7-10} \cmidrule{12-15} \cmidrule{17-20}
    & \multicolumn{1}{c}{\bf best} & \multicolumn{1}{c}{\bf avg} & \multicolumn{1}{c}{\bf s} & \multicolumn{1}{c}{\bf t} & & \multicolumn{1}{c}{\bf best} & \multicolumn{1}{c}{\bf avg} & \multicolumn{1}{c}{\bf s} & \multicolumn{1}{c}{\bf t} & & \multicolumn{1}{c}{\bf best} & \multicolumn{1}{c}{\bf avg} & \multicolumn{1}{c}{\bf s} & \multicolumn{1}{c}{\bf t} & & \multicolumn{1}{c}{\bf best} & \multicolumn{1}{c}{\bf avg} & \multicolumn{1}{c}{\bf s} & \multicolumn{1}{c}{\bf t} \\
    \midrule
    \citet{zhang2017adversarial}, $\lambda = 1$ & 0.00 & 0.00 & 0 & 47.0 & & 0.00 & 0.00 & 0 & 47.0 & & 0.00 & 0.00 & 0 & 45.4 & & 0.00 & 0.00 & 0 & 44.3 \\
     \citet{zhang2017adversarial}, $\lambda = 10$ & 0.00 & 0.00 & 0 & 46.6 & & 0.00 & 0.00 & 0 & 46.0 & & 0.07 & 0.01 & 0 & 44.9 & & 0.07 & 0.01 & 0 & 43.0 \\
     \citet{conneau2018word}, code & 45.40 & 13.55 & 3 & 46.1 & & 47.27 & 42.15 & 9 & 45.4 & & 1.62 & 0.38 & 0 & 44.4 & & 36.20 & 21.23 & 6 & 45.3 \\
     \citet{conneau2018word}, paper & 45.27 & 9.10 & 2 & 45.4 & & 0.07 & 0.01 & 0 & 45.0 & & 0.07 & 0.01 & 0 & 44.7 & & 35.47 & 7.09 & 2 & 44.9 \\
     Proposed method & \bf 48.53 & \bf 48.13 & \bf 10 & \bf 8.9 & & \bf 48.47 & \bf 48.19 & \bf 10 & \bf 7.3 & & \bf 33.50 & \bf 32.63 & \bf 10 & \bf 12.9 & & \bf 37.60 & \bf 37.33 & \bf 10 & \bf 9.1 \\
    \bottomrule
  \end{tabular}
\end{center}
\end{small}
\caption{Results of unsupervised methods on the dataset of \citet{dinu2015improving} and the extensions of \citet{artetxe2017learning,artetxe2018generalizing}. We perform 10 runs for each method and report the best and average accuracies (\%), the number of successful runs (those with \textgreater 5\% accuracy) and the average runtime (minutes).}
\label{tab:results_dinu}
\end{table*}

\section{Experimental settings} \label{sec:experimental_settings}

Following common practice, we evaluate our method on \textbf{bilingual lexicon extraction}, which measures the accuracy of the induced dictionary in comparison to a gold standard.

As discussed before, \textbf{previous evaluation} has focused on favorable conditions. In particular, existing unsupervised methods have almost exclusively been tested on Wikipedia corpora, which is comparable rather than monolingual, exposing a strong cross-lingual signal that is not available in strictly unsupervised settings. In addition to that, some datasets comprise unusually small embeddings, with only 50 dimensions and around 5,000-10,000 vocabulary items \citep{zhang2017adversarial,zhang2017earth}. As the only exception, \citet{conneau2018word} report positive results on the English-Italian dataset of \citet{dinu2015improving} in addition to their main experiments, which are carried out in Wikipedia. While this dataset does use strictly monolingual corpora, it still corresponds to a pair of two relatively close indo-european languages.

In order to get a wider picture of how our method compares to previous work in different conditions, including more challenging settings, we carry out our experiments in the widely used \textbf{dataset} of \citet{dinu2015improving} and the subsequent extensions of \citet{artetxe2017learning,artetxe2018generalizing}, which together comprise English-Italian, English-German, English-Finnish and English-Spanish. More concretely, the dataset consists of 300-dimensional CBOW embeddings trained on WacKy crawling corpora (English, Italian, German), Common Crawl (Finnish) and WMT News Crawl (Spanish). The gold standards were derived from dictionaries built from Europarl word alignments and available at OPUS \citep{tiedemann2012parallel}, split in a test set of 1,500 entries and a training set of 5,000 that we do not use in our experiments. The datasets are freely available. As a non-european agglutinative language, the English-Finnish pair is particularly challenging due to the linguistic distance between them. For completeness, we also test our method in the Spanish-English, Italian-English and Turkish-English datasets of \citet{zhang2017adversarial}, which consist of 50-dimensional CBOW embeddings trained on Wikipedia, as well as gold standard dictionaries\footnote{The test dictionaries were obtained through personal communication with the authors. The rest of the language pairs were left out due to licensing issues.} from Open Multilingual WordNet (Spanish-English and Italian-English) and Google Translate (Turkish-English). The lower dimensionality and comparable corpora make an easier scenario, although it also contains a challenging pair of distant languages (Turkish-English).

Our method is implemented in Python using NumPy and CuPy. Together with it, we also test the \textbf{methods} of \citet{zhang2017adversarial} and \citet{conneau2018word} using the publicly available implementations from the authors\footnote{Despite our efforts, \citet{zhang2017earth} was left out because: 1) it does not create a one-to-one dictionary, thus difficulting direct comparison, 2) it depends on expensive proprietary software 3) its computational cost is orders of magnitude higher (running the experiments would have taken several months).}. Given that \citet{zhang2017adversarial} report using a different value of their hyperparameter $\lambda$ for different language pairs ($\lambda=10$ for English-Turkish and $\lambda=1$ for the rest), we test both values in all our experiments to better understand its effect. In the case of \citet{conneau2018word}, we test both the default hyperparameters in the source code as well as those reported in the paper, with iterative refinement activated in both cases. Given the instability of these methods, we perform 10 runs for each, and report the best and average accuracies, the number of successful runs (those with \textgreater 5\% accuracy) and the average runtime. All the experiments were run in a single Nvidia Titan Xp.

\begin{table*}[t]
\begin{small}
\begin{center}
  \begin{tabular}{clllll}
    \toprule
    \bf Supervision & \bf Method & \bf EN-IT & \bf EN-DE & \bf EN-FI & \bf EN-ES \\
    \midrule
    \multirow{12}{*}{5k dict.} & \citet{mikolov2013exploiting} & 34.93\textsuperscript{$\dagger$} & 35.00\textsuperscript{$\dagger$} & 25.91\textsuperscript{$\dagger$} & 27.73\textsuperscript{$\dagger$} \\
    & \citet{faruqui2014improving} & 38.40\textsuperscript{*} & 37.13\textsuperscript{*} & 27.60\textsuperscript{*} & 26.80\textsuperscript{*} \\  
    & \citet{shigeto2015ridge} & 41.53\textsuperscript{$\dagger$} & 43.07\textsuperscript{$\dagger$} & 31.04\textsuperscript{$\dagger$} & 33.73\textsuperscript{$\dagger$} \\
    & \citet{dinu2015improving} & 37.7 & 38.93\textsuperscript{*} & 29.14\textsuperscript{*} & 30.40\textsuperscript{*} \\
    & \citet{lazaridou2015hubness} & 40.2 & - & - & - \\
    & \citet{xing2015normalized} & 36.87\textsuperscript{$\dagger$} & 41.27\textsuperscript{$\dagger$} & 28.23\textsuperscript{$\dagger$} & 31.20\textsuperscript{$\dagger$} \\
    & \citet{zhang2016ten} & 36.73\textsuperscript{$\dagger$} & 40.80\textsuperscript{$\dagger$} & 28.16\textsuperscript{$\dagger$} & 31.07\textsuperscript{$\dagger$} \\
    & \citet{artetxe2016learning} & 39.27 & 41.87\textsuperscript{*} & 30.62\textsuperscript{*} & 31.40\textsuperscript{*} \\
    & \citet{artetxe2017learning} & 39.67 & 40.87 & 28.72 & - \\
    & \citet{smith2017offline} & 43.1 & 43.33\textsuperscript{$\dagger$} & 29.42\textsuperscript{$\dagger$} & 35.13\textsuperscript{$\dagger$} \\ 
    & \citet{artetxe2018generalizing} & 45.27 & 44.13 & \bf 32.94 & 36.60 \\ 
    \midrule
    25 dict. & \citet{artetxe2017learning} & 37.27 & 39.60 & 28.16 & - \\
    \midrule
    Init. & \citet{smith2017offline}, cognates & 39.9 & - & - & - \\
    heurist. & \citet{artetxe2017learning}, num. & 39.40 & 40.27 & 26.47 & - \\
    \midrule
    \multirow{5}{*}{None}
    & \citet{zhang2017adversarial}, $\lambda = 1$& \phantom{ } 0.00\textsuperscript{*} & \phantom{ } 0.00\textsuperscript{*} & \phantom{ } 0.00\textsuperscript{*} & \phantom{ } 0.00\textsuperscript{*} \\
    & \citet{zhang2017adversarial}, $\lambda = 10$& \phantom{ } 0.00\textsuperscript{*} & \phantom{ } 0.00\textsuperscript{*} & \phantom{ } 0.01\textsuperscript{*} & \phantom{ } 0.01\textsuperscript{*} \\
	& \citet{conneau2018word}, code\textsuperscript{$\ddagger$} & 45.15\textsuperscript{*} & 46.83\textsuperscript{*} & \phantom{ } 0.38\textsuperscript{*} & 35.38\textsuperscript{*} \\
    & \citet{conneau2018word}, paper\textsuperscript{$\ddagger$} & 45.1 & \phantom{ } 0.01\textsuperscript{*} & \phantom{ } 0.01\textsuperscript{*} & 35.44\textsuperscript{*} \\
    & Proposed method & \bf 48.13 & \bf 48.19 & 32.63 & \bf 37.33 \\
    \bottomrule
  \end{tabular}
\end{center}
\end{small}
\caption{Accuracy (\%) of the proposed method in comparison with previous work. \textsuperscript{*}Results obtained with the official implementation from the authors. \textsuperscript{$\dagger$}Results obtained with the framework from \citet{artetxe2018generalizing}. The remaining results were reported in the original papers. For methods that do not require supervision, we report the average accuracy across 10 runs. \textsuperscript{$\ddagger$}For meaningful comparison, runs with \textless 5\% accuracy are excluded when computing the average, but note that, unlike ours, their method often gives a degenerated solution (see Table \ref{tab:results_dinu}).}
\label{tab:results_sota}
\end{table*}

\begin{table*}[t]
\begin{small}
\begin{center}
  \addtolength{\tabcolsep}{-4pt}
  \begin{tabular}{lrrrrrrrrrrrrrrrrrrr}
    \toprule
    & \multicolumn{4}{c}{\bf EN-IT} & & \multicolumn{4}{c}{\bf EN-DE} & & \multicolumn{4}{c}{\bf EN-FI} & & \multicolumn{4}{c}{\bf EN-ES} \\
    \cmidrule{2-5} \cmidrule{7-10} \cmidrule{12-15} \cmidrule{17-20}
    & \multicolumn{1}{c}{\bf best} & \multicolumn{1}{c}{\bf avg} & \multicolumn{1}{c}{\bf s} & \multicolumn{1}{c}{\bf t} & & \multicolumn{1}{c}{\bf best} & \multicolumn{1}{c}{\bf avg} & \multicolumn{1}{c}{\bf s} & \multicolumn{1}{c}{\bf t} & & \multicolumn{1}{c}{\bf best} & \multicolumn{1}{c}{\bf avg} & \multicolumn{1}{c}{\bf s} & \multicolumn{1}{c}{\bf t} & & \multicolumn{1}{c}{\bf best} & \multicolumn{1}{c}{\bf avg} & \multicolumn{1}{c}{\bf s} & \multicolumn{1}{c}{\bf t} \\
    \midrule
    Full system & 48.53 & 48.13 & 10 & 8.9 & & 48.47 & 48.19 & 10 & 7.3 & & 33.50 & 32.63 & 10 & 12.9 & & 37.60 & 37.33 & 10 & 9.1 \\
    \midrule
    - Unsup. init. & 0.07 & 0.02 & 0 & 16.5 & & 0.00 & 0.00 & 0 & 17.3 & & 0.07 & 0.01 & 0 & 13.8 & & 0.13 & 0.02 & 0 & 15.9 \\
    \midrule
    - Stochastic & 48.20 & 48.20 & 10 & 2.7 & & 48.13 & 48.13 & 10 & 2.5 & & 0.28 & 0.28 & 0 & 4.3 & & 37.80 & 37.80 & 10 & 2.6 \\
    - Cutoff ($k$=100k) & 46.87 & 46.46 & 10 & 114.5 & & 48.27 & 48.12 & 10 & 105.3 & & 31.95 & 30.78 & 10 & 162.5 & & 35.47 & 34.88 & 10 & 185.2 \\
    - CSLS & 0.00 & 0.00 & 0 & 15.0 & & 0.00 & 0.00 & 0 & 13.8 & & 0.00 & 0.00 & 0 & 13.1 & & 0.00 & 0.00 & 0 & 14.1 \\
    - Bidirectional & 46.00 & 45.37 & 10 & 5.6 & & 48.27 & 48.03 & 10 & 5.5 & & 31.39 & 24.86 & 8 & 7.8 & & 36.20 & 35.77 & 10 & 7.3 \\
    \midrule
    - Re-weighting & 46.07 & 45.61 & 10 & 8.4 & & 48.13 & 47.41 & 10 & 7.0 & & 32.94 & 31.77 & 10 & 11.2 & & 36.00 & 35.45 & 10 & 9.1 \\
    \bottomrule
  \end{tabular}
\end{center}
\end{small}
\caption{Ablation test on the dataset of \citet{dinu2015improving} and the extensions of \citet{artetxe2017learning,artetxe2018generalizing}. We perform 10 runs for each method and report the best and average accuracies (\%), the number of successful runs (those with \textgreater 5\% accuracy) and the average runtime (minutes).}
\label{tab:results_ablation}
\end{table*}

\section{Results and discussion} \label{sec:results}

We first present the main results (\S \ref{subsec:results_main}), then the comparison to the state-of-the-art (\S \ref{subsec:results_sota}), and finally ablation tests to measure the contribution of each component (\S \ref{subsec:results_ablation}).

\subsection{Main results} \label{subsec:results_main}

We report the results in the dataset of \citet{zhang2017adversarial} at Table \ref{tab:results_zhang}. As it can be seen, the proposed method performs at par with that of \citet{conneau2018word} both in Spanish-English and Italian-English, but gets substantially better results in the more challenging Turkish-English pair. While we are able to reproduce the results reported by \citet{zhang2017adversarial}, their method gets the worst results of all by a large margin. Another disadvantage of that model is that different language pairs require different hyperparameters: $\lambda=1$ works substantially better for Spanish-English and Italian-English, but only $\lambda=10$ works for Turkish-English.

The results for the more challenging dataset from \citet{dinu2015improving} and the extensions of \citet{artetxe2017learning,artetxe2018generalizing} are given in Table \ref{tab:results_dinu}. In this case, our proposed method obtains the best results in all metrics for all the four language pairs tested. The method of \citet{zhang2017adversarial} does not work at all in this more challenging scenario, which is in line with the negative results reported by the authors themselves for similar conditions (only \%2.53 accuracy in their large Gigaword dataset). The method of \citet{conneau2018word} also fails for English-Finnish (only 1.62\% in the best run), although it is able to get positive results in some runs for the rest of language pairs. Between the two configurations tested, the default hyperparameters in the code show a more stable behavior.

These results confirm the robustness of the proposed method. While the other systems succeed in some runs and fail in others, our method converges to a good solution in all runs without exception and, in fact, it is the only one getting positive results for English-Finnish. In addition to being more robust, our method also obtains substantially better accuracies, surpassing previous methods by at least 1-3 points in all but the easiest pairs. Moreover, our method is not sensitive to hyperparameters that are difficult to tune without a development set, which is critical in realistic unsupervised conditions.

At the same time, our method is significantly faster than the rest. In relation to that, it is interesting that, while previous methods perform a fixed number of iterations and take practically the same time for all the different language pairs,
the runtime of our method adapts to the difficulty of the task thanks to the dynamic convergence criterion of our stochastic approach. This way, our method tends to take longer for more challenging language pairs (1.7 vs 0.6 minutes for es-en and tr-en in one dataset, and 12.9 vs 7.3 minutes for en-fi and en-de in the other) and, in fact, our (relative) execution times correlate surprisingly well with the linguistic distance with English (closest/fastest is German, followed by Italian/Spanish, followed by Turkish/Finnish).

\subsection{Comparison with the state-of-the-art} \label{subsec:results_sota}

Table \ref{tab:results_sota} shows the results of the proposed method in comparison to previous systems, including those with different degrees of supervision. We focus on the widely used English-Italian dataset of \citet{dinu2015improving} and its extensions.  Despite being fully unsupervised, our method achieves the best results in all language pairs but one, even surpassing previous supervised approaches. The only exception is English-Finnish, where \citet{artetxe2018generalizing} gets marginally better results with a difference of 0.3 points, yet ours is the only unsupervised system that works for this pair. At the same time, it is remarkable that the proposed system gets substantially better results than \citet{artetxe2017learning}, the only other system based on self-learning, with the additional advantage of being fully unsupervised.

\subsection{Ablation test} \label{subsec:results_ablation}

In order to better understand the role of different aspects in the proposed system, we perform an ablation test, where we separately analyze the effect of initialization, the different components of our robust self-learning algorithm, and the final symmetric re-weighting. The obtained results are reported in Table \ref{tab:results_ablation}.

In concordance with previous work, our results show that self-learning does not work with random initialization. However, the proposed unsupervised initialization is able to overcome this issue without the need of any additional information, performing at par with other character-level heuristics that we tested (e.g. shared numerals).

As for the different self-learning components, we observe that the stochastic dictionary induction is necessary to overcome the problem of poor local optima for English-Finnish, although it does not make any difference for the rest of easier language pairs. The frequency-based vocabulary cutoff also has a positive effect, yielding to slightly better accuracies and much faster runtimes. At the same time, CSLS plays a critical role in the system, as hubness severely accentuates the problem of local optima in its absence. The bidirectional dictionary induction is also beneficial, contributing to the robustness of the system as shown by English-Finnish and yielding to better accuracies in all cases.

Finally, these results also show that symmetric re-weighting contributes positively, bringing an improvement of around 1-2 points without any cost in the execution time.

\section{Conclusions} \label{sec:conclusions}

In this paper, we show that previous unsupervised mapping methods \cite{zhang2017adversarial,conneau2018word} often fail on realistic scenarios involving non-comparable corpora and/or distant languages. In contrast to adversarial methods, we propose to use an initial weak mapping that exploits the structure of the embedding spaces in combination with a robust self-learning approach. The results show that our method succeeds in all cases, providing the best results with respect to all previous work on unsupervised and supervised mappings.

The ablation analysis shows that our initial solution is instrumental for making self-learning work without supervision. In order to make self-learning robust, we also added stochasticity to dictionary induction, used CSLS instead of nearest neighbor, and produced bidirectional dictionaries. Results also improved using smaller intermediate vocabularies and re-weighting the final solution. Our implementation is available as an open source project at \url{https://github.com/artetxem/vecmap}.

In the future, we would like to extend the method from the bilingual to the multilingual scenario, and go beyond the word level by incorporating embeddings of longer phrases.

\section*{Acknowledgments}

This research was partially supported by the Spanish MINECO (TUNER TIN2015-65308-C5-1-R, MUSTER PCIN-2015-226 and TADEEP TIN2015-70214-P, cofunded by EU FEDER), the UPV/EHU  (excellence research group), and the NVIDIA GPU grant program. Mikel Artetxe enjoys a doctoral grant from the Spanish MECD.


\bibliography{acl2018}
\bibliographystyle{acl_natbib}

\end{document}